\documentclass{article}
\usepackage[preprint]{spconf}
\usepackage{spconf,amsmath,graphicx}
\usepackage{amsfonts, calrsfs}
\usepackage{graphicx}
\usepackage{subcaption}
\usepackage[numbers,sort,compress]{natbib}
\usepackage{enumitem}
\usepackage{enumerate}

\title{Loss Switching Fusion with Similarity Search for Video Classification}
\name{Lei Wang$^{\star \dagger} \sthanks{The author performed the work while he was a Computer Vision Research Intern at iCetana Pty Ltd.} $ \qquad Du Q.~Huynh $^{\dagger}$ \qquad Moussa Reda Mansour $^{\star \dagger}$  }
\address{$^{\star}$ iCetana Pty Ltd\\  Suite 4/6 Centro Ave, Subiaco WA 6008
\\ $^{\dagger}$ Department of Computer Science and Software Engineering \\ The University of Western Australia \\ Mounts Bay Road, Crawley WA 6009}

\newcommand{\Np}{N_{\text{p}}}
\newcommand{\Npp}{N'_{\text{p}}}
\newcommand{\LSFNet}{{\mbox{LSFNet}}}

\begin{document}
%
\maketitle

\copyrightnotice{
\copyright\ IEEE 2019}\toappear{To appear in {\it Proc.\ ICIP2019, September 22-25, 2019, Taipei, Taiwan}}
\begin{abstract}

From video streaming to security and surveillance applications, video data play an important role in our daily living today. However, managing a large amount of video data and retrieving the most useful information
for the user remain a challenging task. 
In this paper, we propose a novel video classification system that would benefit the scene understanding task. We define our classification problem as classifying background and foreground motions using the same feature representation for outdoor scenes. This means that the feature representation needs to be robust enough and adaptable to different classification tasks.
We propose a lightweight Loss Switching Fusion Network (\LSFNet) for the fusion of spatiotemporal descriptors and a similarity search scheme with soft voting to boost the classification performance. The proposed system has a variety of potential applications such as content-based video clustering, video filtering, etc. Evaluation results on two private industry datasets show that our system is robust in both classifying different background motions 
and detecting human motions from these background motions. 

\end{abstract}
\section{Introduction}
\label{sec:intro}


Scene understanding~\cite{greg2014, mark2016, arunnehru2018, mateus2018, mehrsan2013, waqas2018} in noisy videos is a challenging task. Firstly, videos contain too much redundant information, which slows down the training process and makes the feature extraction much harder.
Secondly, 
some feature extraction techniques~\cite{geert2008, herbert2006, navneet2006} ignore the relationship between spatial and temporal information as they extract such information separately, which leads to the loss of information. 
Thirdly, scene understanding needs to fuse both background and foreground motions to represent the information within a given video~\cite{chen2018, arunnehru2018}, and this is a challenging research area that has not been fully explored.

Content-based video clustering and classification~\cite{mohamed2013, feifei2005, kaiqi2011} can significantly increase the speed of tasks like searching and browsing for a particular video, and with the increasing need for security and retrieval, this kind of system is of vital importance to predict and avoid unwanted scenes or behaviours.
Compared to indoor scenes~\cite{Rahmani2016, RahmaniHOPC2014}, video classification in outdoor scenes~\cite{mitko2009, limin2018, haichen2017, xiang2018} are much harder as there are many dynamic environment motions such as raining and tree waving that can affect the performance of classification. Although neural network techniques have achieved great success in many fields~\cite{du2015, joao2018, katsunori2016, limin2015, kensho2017}, they require a lot of training data which are not easy to obtain from industry. Moreover, the industry data are far more complex than the benchmarks used in research due to the dynamic changes of data distributions over time. Compared to the neural network methods, feature engineering~\cite{uijlings2014, heng2011, alexander2008, paul2007, heng2013, heng2013iccv} enjoys the flexibility, computational efficiency, and does not rely on large sets of samples for training.

We propose a novel video classification system for background and foreground motion classification that combines the merits of both neural network and feature engineering for industry applications. We define our video classification problem as classifying videos based on dynamic environment motions such as tree waving, noise, and camera shaking, and detecting human motions from these dynamic environment motions using the same feature representation for outdoor scenes. 
This leads to a better scene understanding system as recomputing features for different tasks 
is time consuming and computationally expensive. We refer to dynamic environment motions and human motions as background and foreground motions respectively from hereon.
Our research contributions are:
\begin{itemize}[topsep=0pt,itemsep=-2pt]
    \item We propose a novel video classification system for video clustering, focusing on background and foreground motion using the same feature representation.
    \item We introduce a lightweight fusion network to fuse spatiotemporal features non-linearly, based on the concept of `loss switching'.
    \item 
    We propose to use similarity search with soft voting for robust video classification. 
\end{itemize}

\begin{figure}[tb!]
\centering
\includegraphics[width=0.9\linewidth]{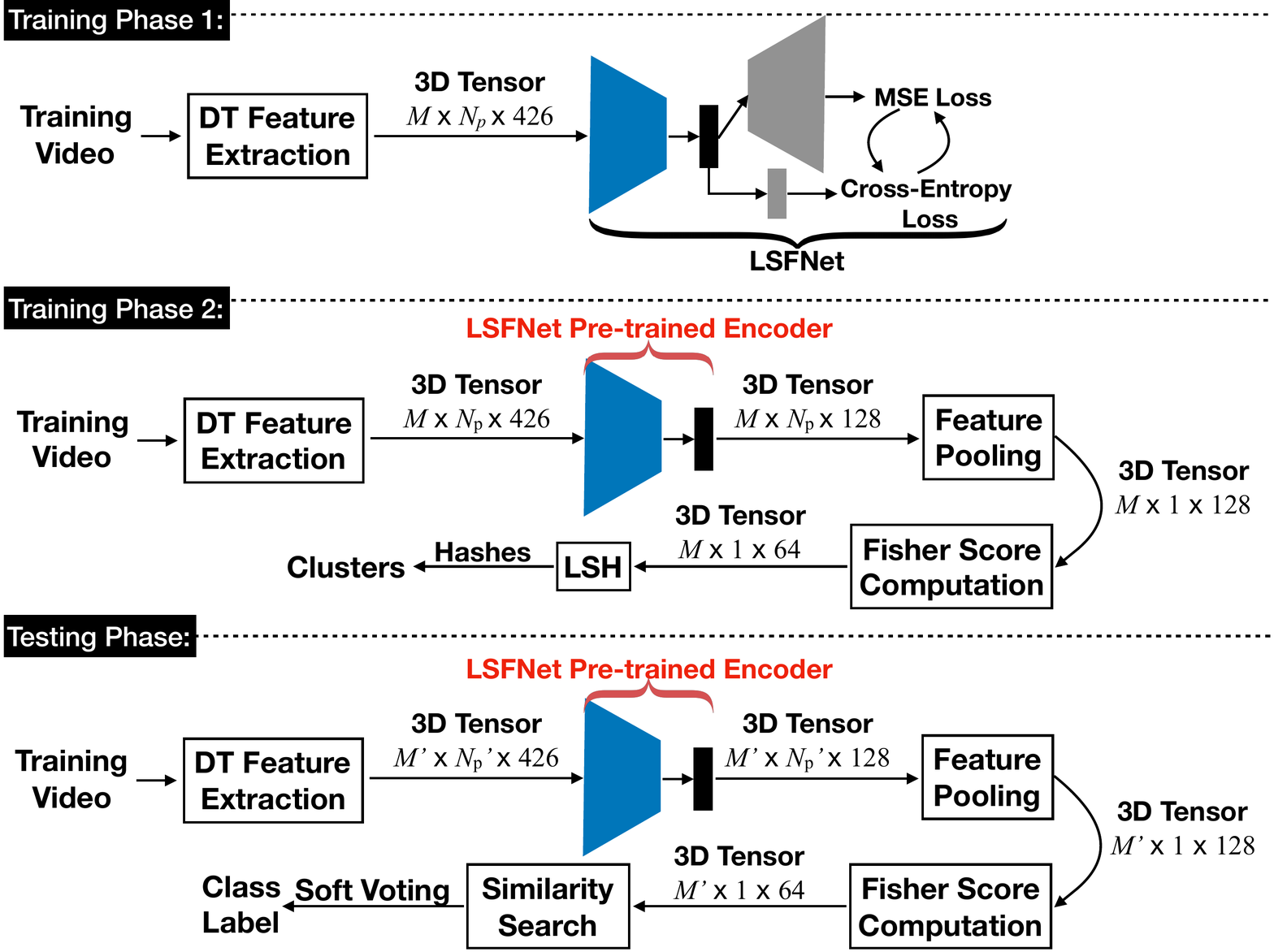}
\\
\caption{The training and testing stages of the proposed system.
$M$ and $M'$ are the number of training and testing samples respectively. $\Np$ and $\Npp$ are the number of tracked motion points in training and testing videos respectively, and they are variable depending on the motion trajectories of the videos.
} 
\label{system}
\end{figure}

\section{Method}
\label{sec:method}

Figure~\ref{system} illustrates the basic work flow of our proposed approach. 
There are two training phases.
In phase 1, a lightweight Loss Switching Fusion Network (\LSFNet) is trained using spatiotemporal descriptors randomly sampled from the videos as input. In phase 2, the pre-trained \LSFNet~and a feature pooling layer together output a lower-dimensional feature vector for representing each video. 
We then compute the Fisher scores to rank and select the most discriminative features to enhance the feature separability.
To efficiently store and retrieve these videos for later use, we adopt a Locality Sensitive Hashing  (LSH)~\cite{li2015, wang2017} mechanism, which returns similar hashes for similar videos.
This process maps the training videos into several classes based on their feature distances. So in addition to video retrieval, the system also facilitates video classification.
In the testing phase, for each test video, we use similarity search to find the most similar feature representations so as to get their corresponding labels. After that, we count and compare the number of labels retrieved using `soft voting' to get the confidence values to assign label to each test video.  



\subsection{Spatiotemporal Features}
\label{sec:st-features}
Dense trajectories (DT)~\cite{heng2011, heng2013} and improved trajectories (iDT)~\cite{heng2013iccv} are the state-of-the-art handcrafted features that have achieved great success in human action recognition due to its robustness in mining motion trajectories. The DT features are 
normalized
trajectory-aligned descriptors 
comprising spatiotemporal HOG (ST-HOG), spatiotemporal HOF (ST-HOF), spatiotemporal MBH (ST-MBH) and spatiotemporal trajectories (ST-DT). 
Our proposed system uses these four spatiotemporal features as the base features to describe the videos. 



\begin{figure}[t!]
\centering
\includegraphics[width=0.75\linewidth]{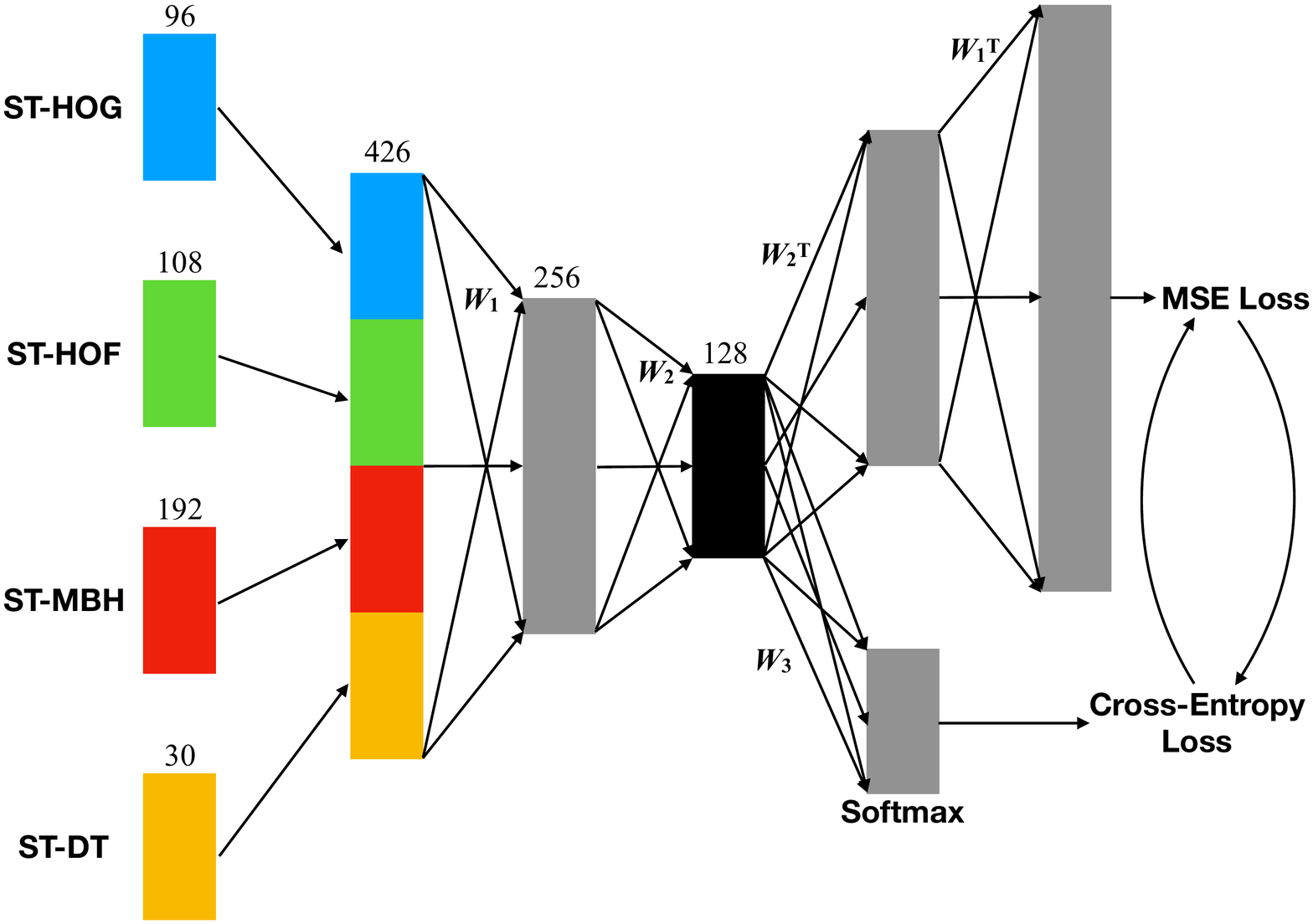}\\
\caption{The proposed \LSFNet~for the fusion of spatiotemporal descriptors.
} 
\label{fusionnet}
\end{figure}

\subsection{Loss Switching Fusion Network (\LSFNet)}

Many methods such as Principal Component Analysis (PCA) and traditional autoencoder have been used for feature dimensionality reduction. While the PCA focuses on analyzing the covariance matrix of the features, the traditional autoencoder learns a transformation that minimizes the reconstruction error. Both methods ignore the fact that the feature data may form clusters and, by retaining this cluster information in the dimension reduction process, the lower dimensional features would be more useful for the downstream feature classification process. 


To retain the underlying cluster structure of the spatiotemporal features that describe each video, rather than just concatenating the features and passing the resultant long vectors to the PCA or autoencoder, we propose to fuse the spatiotemporal features nonlinearly by training a lightweight \LSFNet~shown in Figure~\ref{fusionnet}. 
The inputs to the network are the four
normalized trajectory-aligned spatiotemporal descriptors described in Section~\ref{sec:st-features}.
Our \LSFNet~is composed of two small sub-networks: (i)~a 5-layer autoencoder having weights symmetrically tied around the middle encoding layer; (ii)~a multi-layer perceptron (MLP) classifier, which shares the encoder part of the autoencoder. 
There are two separate loss functions in \LSFNet:
\begin{itemize}[topsep=0pt,itemsep=0pt]
\item The first loss function, denoted by $L_1$, is the MSE loss for the autoencoder to  minimize the reconstruction errors of features.

\item The second loss function, denoted by $L_2$, is the cross-entropy loss for measuring the classification performance.
\end{itemize}
In training phase 1 (see Fig.~\ref{system}), 
the MSE loss $L_1$ and classification loss $L_2$ are used alternately in each pass of the gradient descent. The role of the MLP classifier is to steer the optimization so that the reduced-dimensional feature vectors produced in the encoding layer are more separated (or closer together) if they belong to different classes (or the same class). In training phase 2 and the testing phase, only the part up to the encoding layer is used.



\subsection{Feature Selection}

\newcommand{\bx}{\boldsymbol x}

After feature fusion, the feature dimension of each video is $\Np \times 128$, where $\Np$ is the number of tracked motion points within a 15-frame subsequence~\cite{heng2011}. Long and/or complex videos (containing many motion trajectory points) would have large $\Np$ value. The average pooling layer of the system (Fig.~\ref{system}) then reduces the feature 
down to $1 \times 128$ to get a holistic representation for each video. The next component of the pipeline takes care of the Fisher score~\cite{arockiam2012, sa2007} computation which ranks the feature importance to yield a discriminative feature that has a more compact representation. The Fisher score of the $i^{\text{th}}$ feature component is given by: 
\begin{equation}
    f_i
    = \textstyle\sum_{c = 1}^C n_c(\mu_c^i - \mu^i)^2/\textstyle\sum_{c = 1}^C n_c ( \sigma_c^i)^2
\end{equation}
where $\mu_c^i$ and $\sigma_c^j$ are, respectively, the mean and standard deviation of the $c^{\text{th}}$ class for the $i^{\text{th}}$ feature component, $\mu^i$ denotes the mean of the whole dataset corresponding to the $i^{\text{th}}$ feature, and $n_c$ is the size of the $c^{\text{th}}$ class. We then rank $f_i$ in descending order, and get the indexes of the feature components that have the top-$q$\% highest Fisher scores. Based on our experiments and the trade-off between the computational cost and classification accuracy, we choose $q = 50$ so that the feature dimension is $1 \times 64$ after feature selection.  
In the testing phase after feature pooling, we select the feature components using the feature indexes obtained from the training stage for later processing described in Section~\ref{sec:lsh}.




\subsection{Similarity Search for Classification}
\label{sec:lsh}
LSH~\cite{li2015}  is defined based on the simple idea that, if two points are close together, then after a projection operation, these two points will remain close together. So LSH targets at mapping the 
features in such a way that similar features have a high probability to be mapped to similar index values. 
To quickly map a given feature vector to a hash value, the scalar projection given in~\cite{wang2017} is commonly used as the hash function.
Let $\bx$ be the 64D feature vector (representing a video) obtained from the feature selection process above, we use this scalar projection hash function, denoted by $h$, to map $\bx$ to a hash value in $\mathbb{R}$. To generate more accurate video clusters, we adopt 
$N$ different such hash functions. Let 
$
    g(\boldsymbol x) = [h_1(\boldsymbol x), h_2(\boldsymbol x), \cdots, h_N(\boldsymbol x)] \in \mathbb{R}^N.
$ 
For a $C$-class classification problem, $C$ clusters would be created and $C$ such $g$ functions would be used:
$g_1, g_2, \cdots, g_C$. By having multiple hash functions for multiple clusters, it helps to improve the accuracy of feature clustering if each query feature and its nearest neighbours fall in the same cluster in all the $N$ scalar projections.

In the testing phase, given a query feature $\boldsymbol y$ computed by the feature selection stage for a test video, the system computes 
$g_c(\boldsymbol y)$, for $c = 1,\cdots,C$. 
After that, 
it performs a linear search to find $K$ nearest neighbours of the query feature in the $\mathbb{R}^N$ space, 
where $K$ is a value defined by the user.
We then use `soft voting' to count and compute the confidence value of assigning the query feature to each cluster. 


\section{Datasets and Experimental Settings}
\label{sec:experiment}
\subsection{Industry Datasets}
We use two industry datasets to evaluate the clustering and classification performance of the proposed system:

%
%
$\bullet$ {\bf iCetanaPrivateDataset}. 
This dataset  contains 2700 videos of various lengths. They were captured in outdoor environments so issues, such as tree waving, camera shaking, noise, illumination changes, and rain, are common. Some videos have human motion also. The outdoor scenes include car parks, train stations, bus stops, etc. 

$\bullet$ {\bf iCetanaEventDataset}. This dataset is an extension of iCetanaPrivateDataset, with videos captured by multiple cameras located at different train stations, bus stops, moving lifts, restaurants, and supermarkets. It has 6668 videos. 

The average length of the videos in both datasets is $\sim$280 frames, with long videos up to 19645 frames. 
Both datasets have been manually labelled with 6 background motion class labels: \textit{tree waving}, \textit{camera shaking}, \textit{noisy video}, \textit{rainy}, \textit{illumination}, and \textit{normal video}. There are 3 foreground human motion class labels: \textit{general body movements}, \textit{human-object interaction}, and \textit{human-human interaction}. Videos having both background and foreground motions have two class labels. In this paper, the 3 types of foreground motion are grouped together into one class.





\subsection{Experimental Settings}
We evaluate the performance of our proposed system by 
\begin{itemize}[topsep=0pt,itemsep=-2pt]
\item testing how well the 6 background motion classes are classified. This is a multi-class classification problem;
\item testing how well the foreground motion is separated from those background motions. 
This is a binary classification problem.
\end{itemize}

To extract the trajectory-aligned descriptor from video, we use the codes from~\cite{heng2011}. For the feature fusion part in training phase 1, $10^6$ 
trajectory-aligned descriptors were randomly sampled from iCetanaEventDataset along with their background motion class labels 
are passed to the network. The learning rates for training the autoencoder and MLP were set to $10^{-3}$ and $10^{-2}$ respectively. The network was trained using a MacBook Pro 2.9 GHz Intel Core i7 computer for up to 200 iterations (this was shown to be sufficient due to the large number of training samples). The whole training process took around half a day to finish. We also compare the performance of our fusion with PCA and autoencoder alone by truncating the features down to the same dimension (128D) as \LSFNet. For PCA, this leads to a loss of about 9\% feature importance. For autoencoder alone, the learning rate was set to $10^{-1}$ with a learning rate decay of $10^{-1}$ after every 50 epochs and the number of iterations was also set to 200.

For both training and testing, we randomly choose $N$ (number of hash functions for LSH) from the range $[10, 50]$ and select $K$ (number of nearest neighbouring videos for similarity search) from $[50, 100]$. The performance for each classification task is computed by averaging over different $(N, K)$ pairs.



To compare with the state-of-the-art techniques, we modified the last prediction layer of I3D~\cite{joao2018} (or C3D~\cite{du2015}) to have 6 background motion classes and passed the videos from the iCetanaEventDataset to fine-tune the last Inception-v1 module (or the last fully connected layer) and then tested the classification performance on iCetanaPrivateDataset.




\section{Results and Discussions}
\label{sec:result}

\begin{figure}[tb!]
    \centering 
\begin{subfigure}{0.27\textwidth}
  \includegraphics[width=\linewidth]{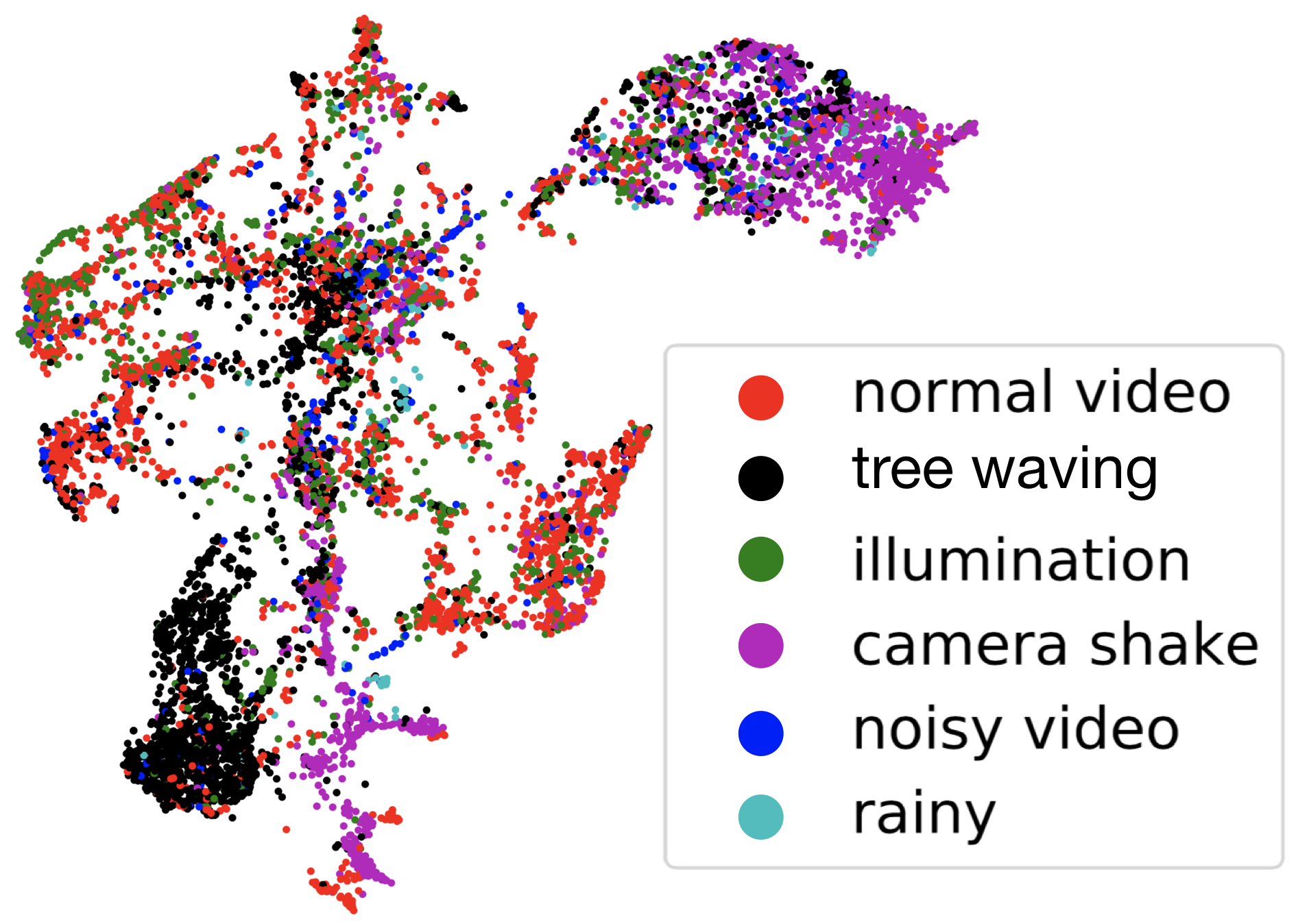}
  \caption{Simple concat. + PCA}
  \label{concate_env}
\end{subfigure}\hfil 
\begin{subfigure}{0.21\textwidth}
  \includegraphics[width=\linewidth]{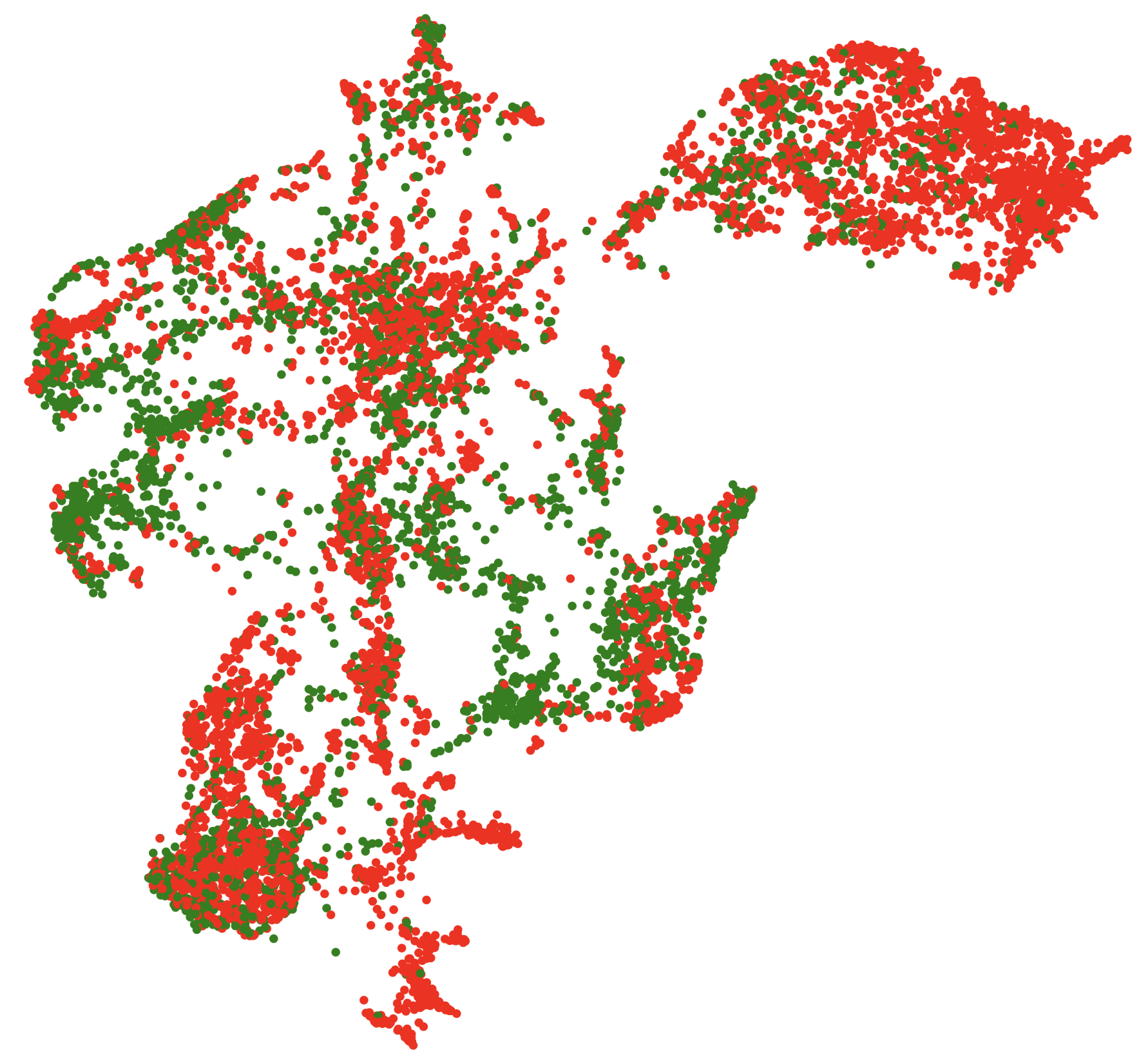}
  \caption{Simple concat. + PCA}
  \label{concate_human}
\end{subfigure}
\begin{subfigure}{0.24\textwidth}
  \includegraphics[width=\linewidth]{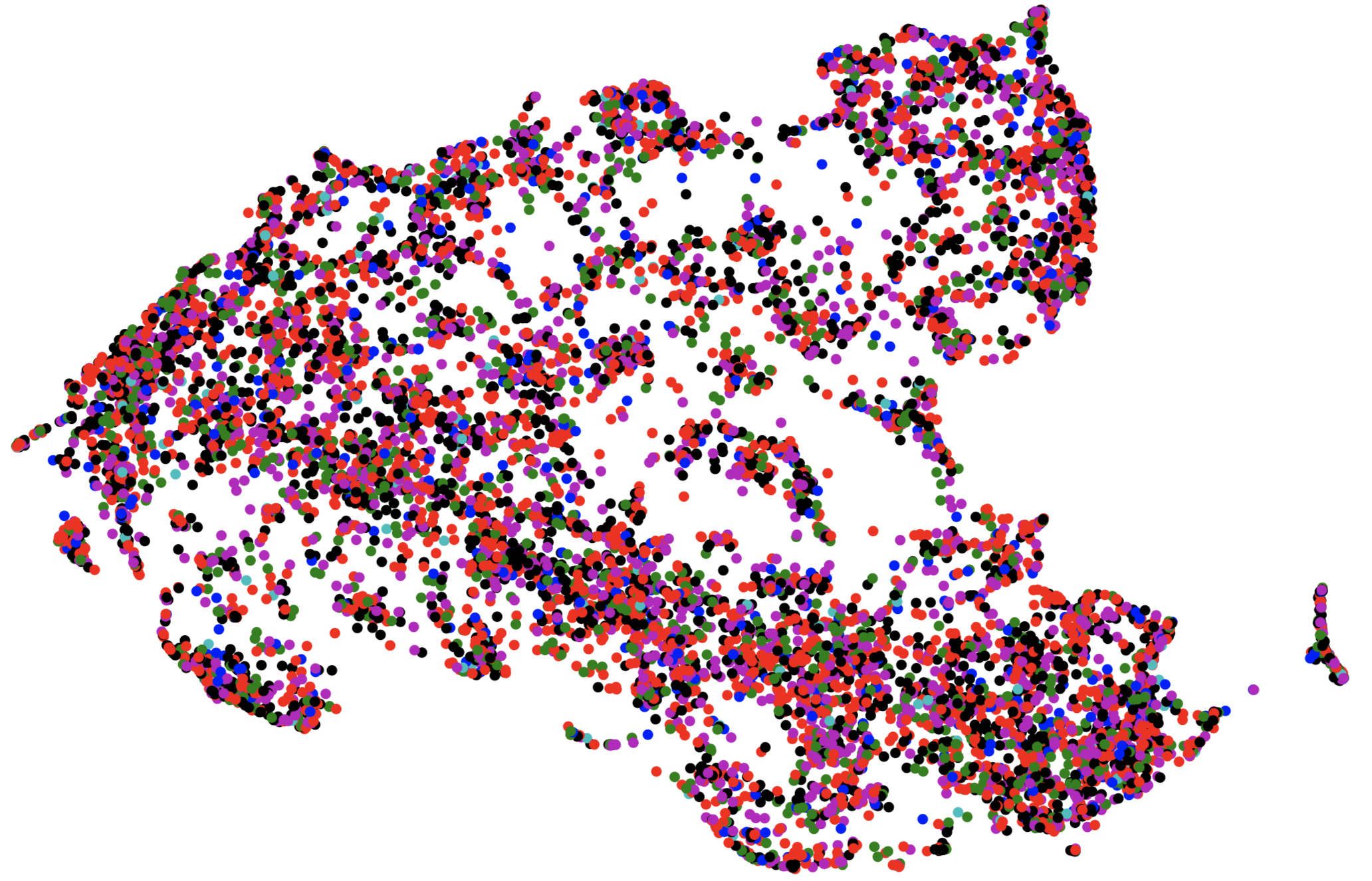}
  \caption{Autoencoder alone}
  \label{no_mlp_env}
\end{subfigure}\hfil 
\begin{subfigure}{0.24\textwidth}
  \includegraphics[width=\linewidth]{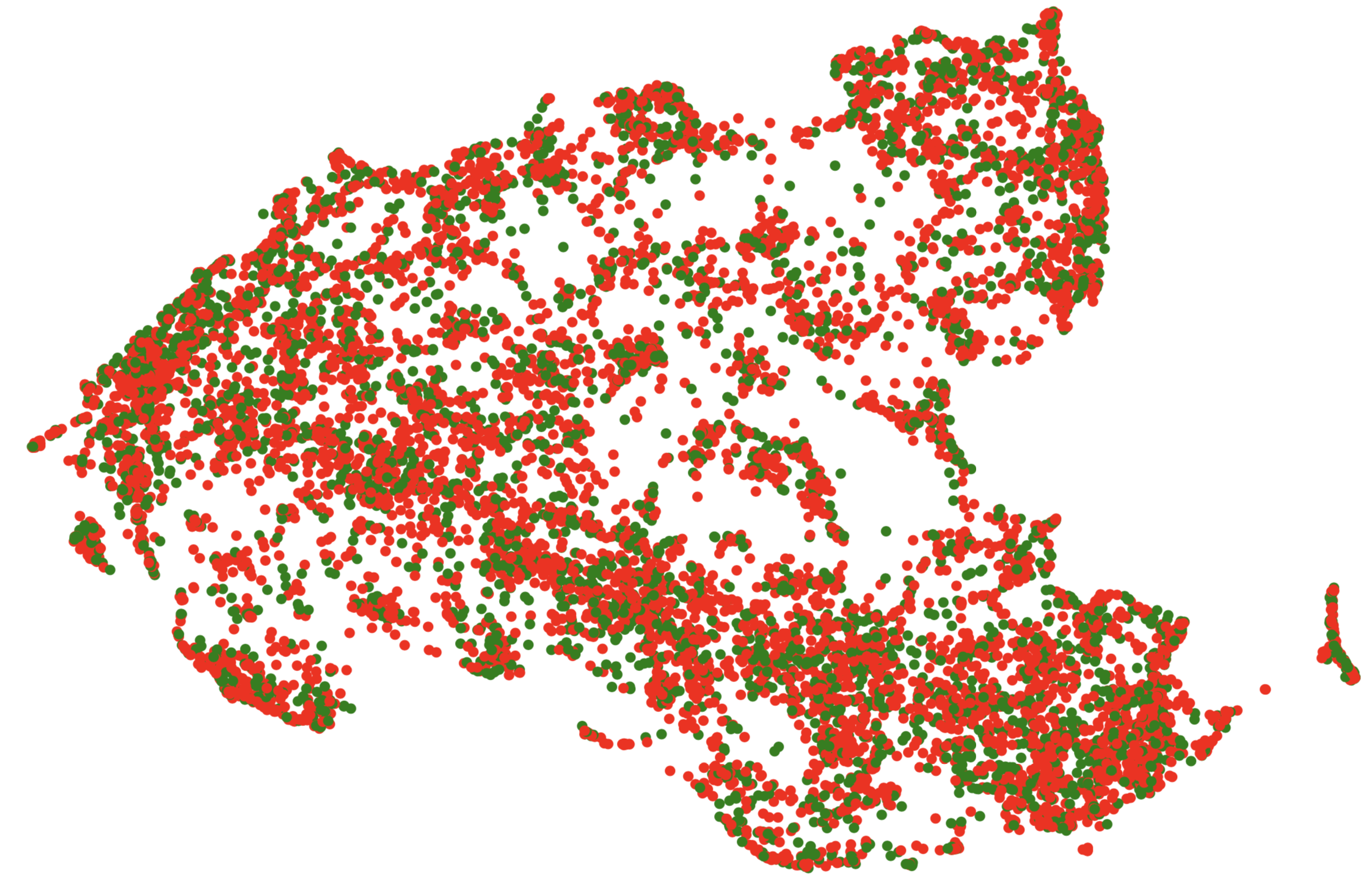}
  \caption{Autoencoder alone}
  \label{no_mlp_human}
\end{subfigure}
\begin{subfigure}{0.24\textwidth}
  \includegraphics[width=\linewidth]{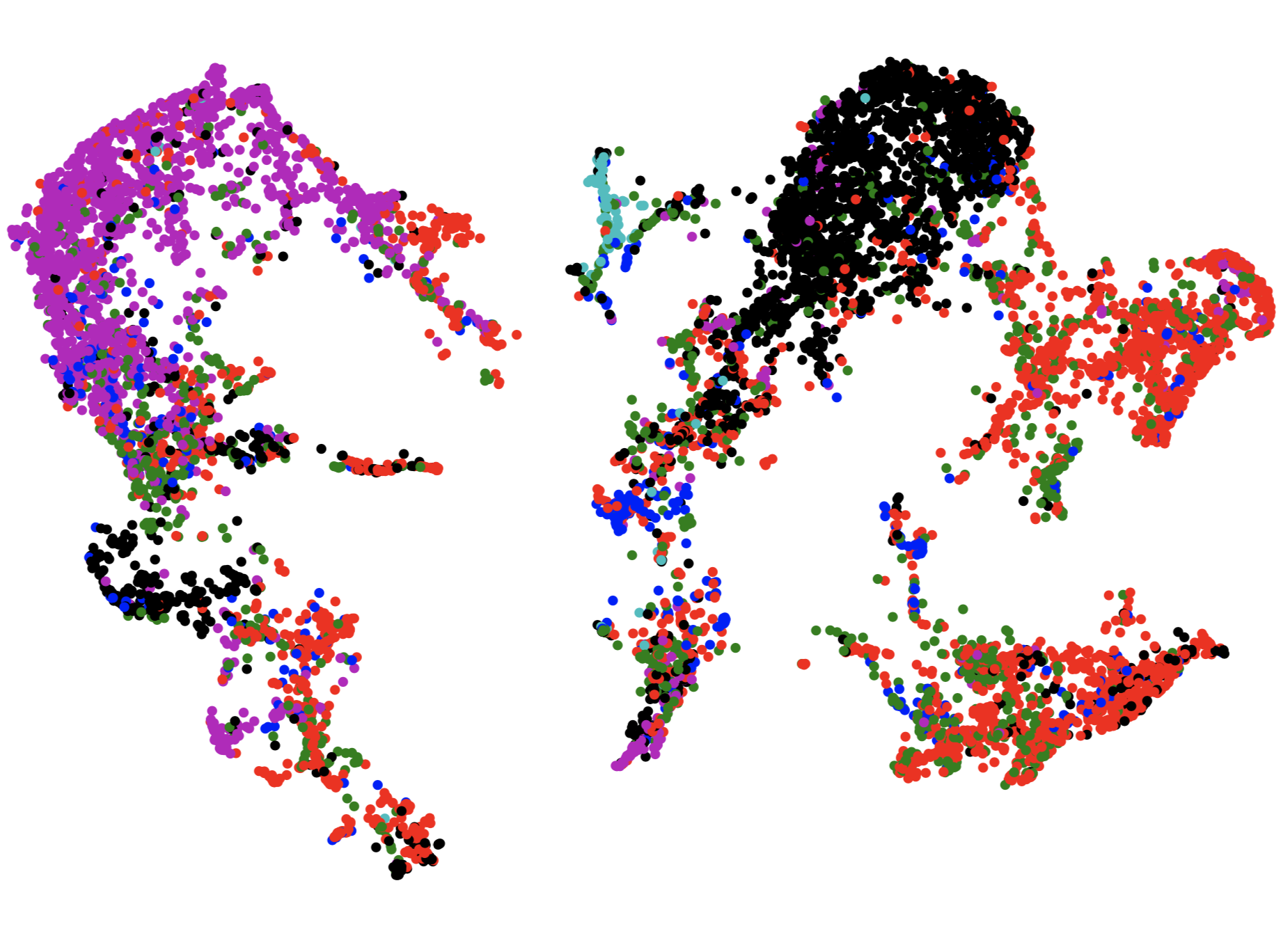}
  \caption{\LSFNet~fusion}
  \label{lsf_env}
\end{subfigure}\hfil 
\begin{subfigure}{0.24\textwidth}
  \includegraphics[width=\linewidth]{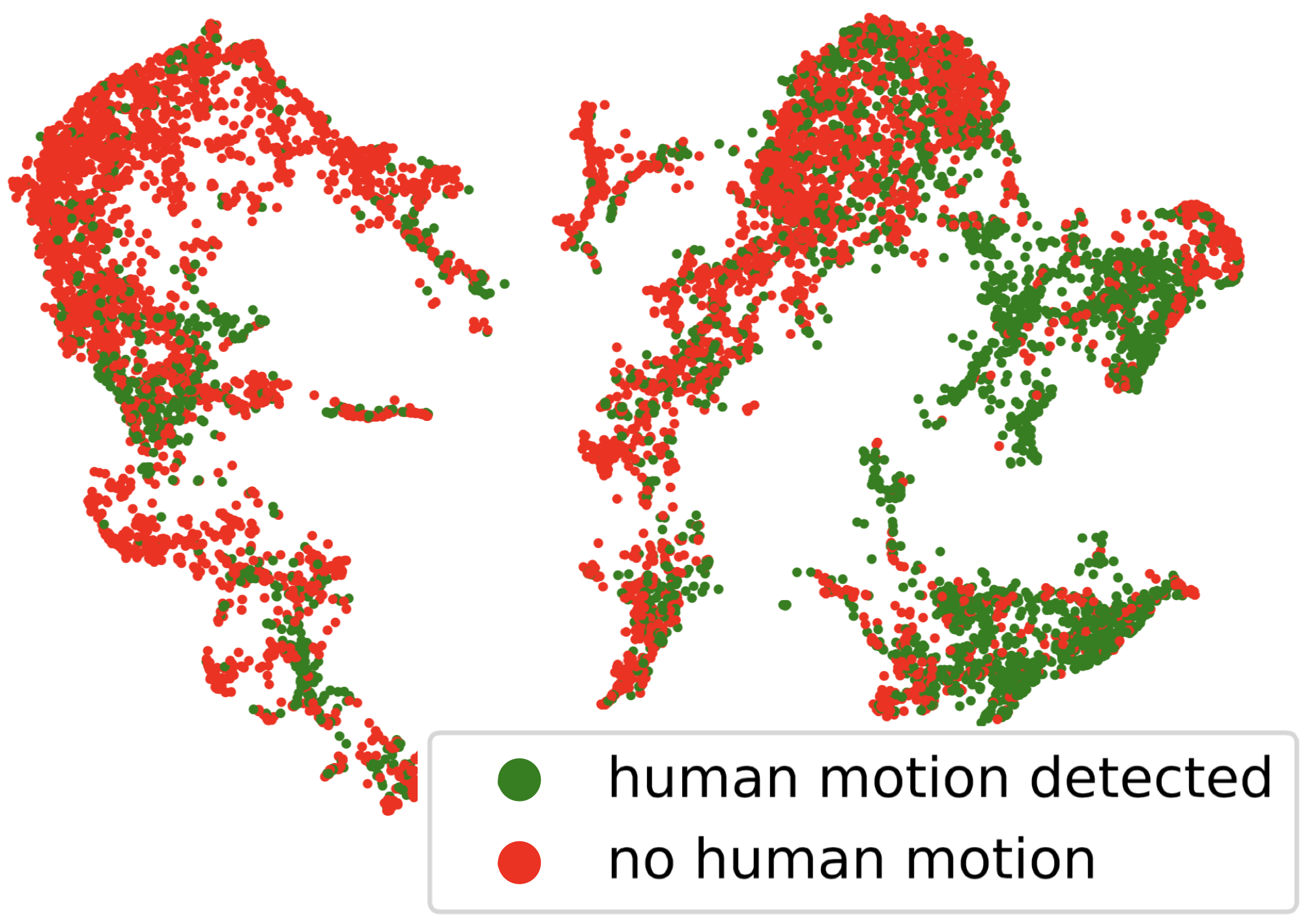}
  \caption{\LSFNet~fusion}
  \label{lsf_human}
\end{subfigure}
    \caption{Feature space visualization for background and foreground motions on the testing set of the iCetanaPrivateDataset using simple concatenation of spatiotemporal descriptors followed by PCA, standard autoencoder and our \LSFNet~fusion. 
    The figures in each column share the same legend.
    }
    \label{cluster_comp}
\end{figure}

\textbf{Video Clustering. }
We evaluate our video clustering performance using the reduced-dimensional features from \LSFNet~against those from the PCA and standard autoencoder.  Figure~\ref{cluster_comp} shows the comparison results. The visualization was produced using Uniform Manifold Approximation and Projection (UMAP)~\cite{mcinnes2018umap-software, 2018arXivUMAP}. Comparing the results from using autoencoder alone, it is clear that, for both background and foreground motion classes, \LSFNet~produces more compact clusters for videos of the same motion class and more separated clusters for videos of different motion classes. While the results from PCA are better than those from using autoencoder alone, there is some intertwining among some motion classes.

\begin{table}[t!]
\caption{A comparison with state-of-the-art methods for background and foreground motion classification.}
\vspace{-0.5cm}
\begin{center}
\resizebox{0.48\textwidth}{!}{\begin{tabular}{| l | c | c |}
\hline
 & Background & Foreground \\
 Algorithms & {env. motion} & {human motion} \\ 
\hline
\hline
iDT~\cite{heng2013iccv}  & 48.1 & 66.7\\
\hline
C3D~\cite{du2015} (Sports 1M pre-training) + LinearSVM & 74.1 & 70.4 \\
\hline

C3D~\cite{du2015} (finetuned using iCetanaEventDataset) & 75.9 & 77.8\\
\hline
I3D RGB~\cite{joao2018}(finetuned using iCetanaEventDataset) & 77.0  & 79.9 \\
\hline
\hline
$\text{Fisher score + CCA}^{\dagger}$  & 81.5  & 85.2 \\
\hline
$\text{DT + FV + Fisher score + LSH}^{\ddagger}$  & 83.8  & 86.5 \\
\hline
\LSFNet & 83.3  & 85.2 \\
\hline
\LSFNet + Fisher score & 85.2  & 87.0 \\
\hline
{Our whole system} & {\bf 88.9} & {\bf 90.7} \\
\hline 
\multicolumn{3}{l}{\parbox{1.5\linewidth}{
$^\dagger$Our own pipeline using Fisher score for each spatiotemporal descriptor followed by Canonical Correlation Analysis (CCA)~\cite{arunnehru2018} for the feature fusion. \\
$^\ddagger$Our own pipeline using DT~\cite{heng2011} followed by Fisher vector (FV)~\cite{florent2007, florent2010}, then Fisher score is used to select the top-50\% feature components for LSH.}}
\end{tabular}}
\label{overallresults}
\end{center}
\vspace{-1ex}
\end{table}

\noindent \textbf{Video Classification.}
Table~\ref{overallresults} shows a comparison of our method with other state-of-the-art techniques. As shown in the table, our whole system achieves the best among all other techniques in both background and foreground motion classification. With \LSFNet, the performance is better than fine-tuning the deep learning models such as C3D and I3D. This improvement is gained from the loss switching mechanism which forces the learned representation to be more discriminative and compact. It should also be noted that \LSFNet~performs better than CCA~\cite{arunnehru2018}, which is the feature fusion technique using correlation analysis. With LSH, the classification performance increases by 4\%.

\section{Conclusion}
\label{sec:conclusion}

We have presented a lightweight video classification system that can robustly classify videos that have background and foreground motions. 
There are also additional types of background motions like snowing, thunder storm and fogging, we leave them for future work. We also introduce an \LSFNet~for the fusion of spatiotemporal descriptors and our method achieves the best clustering and classification results compared to existing techniques, including some complex deep learning models. 
Our future work will focus on exploring more powerful feature fusion pipelines suitable for mid-level and high-level feature fusion. 



\scriptsize{\bibliographystyle{IEEEbib}
\bibliography{refs}}

\begin{thebibliography}{10}

\bibitem{greg2014}
Greg castanon, Mohamed Elgharib, Venkatesh Saligrama, and Pierre-Marc Jodoin,
\newblock ``{Retrieval in Long Surveillance Videos using User Described Motion
  and Object Attributes},''
\newblock {\em IEEE Transactions on Multimedia}, pp. 1--13, 2014.

\bibitem{mark2016}
Mark Marsden, Kevin McGuinness, Suzanne Little, and Noel~E. O'Connor,
\newblock ``{Holistic Features for Real-time Crowd Behaviour Anomaly
  Detection},''
\newblock {\em ICIP}, 2016.

\bibitem{arunnehru2018}
J.~Arunnehru, A.~Yashwanth, and Shaik Shammer,
\newblock ``{Canonical Correlation-Based Feature Fusion Approach for Scene
  Classification},''
\newblock {\em International Conference on Intelligent Systems Design and
  Applications}, pp. 134--143, 2018.

\bibitem{mateus2018}
Mateus~T. Nakahata, Lucas~A. Thomaz, and Allan~F. da~Silva,
\newblock ``{Anomaly detection with a moving Camera using Spatio-temporal
  Codebooks},''
\newblock {\em Multidim Syst Sign Process}, pp. 1025--1054, 2018.

\bibitem{mehrsan2013}
Mehrsan~Javan Roshtkhari and Martin~D. Levine,
\newblock ``{An Online, Realtime Learning Method for Detecting Anomalies in
  Video using Spatio-temporal Compositions},''
\newblock {\em CVIU}, 2013.

\bibitem{waqas2018}
Waqas Sultani, Chen Chen, and Mubarak Shah,
\newblock ``{Real-world Anomaly Detection in Surveillance Videos},''
\newblock {\em CVPR}, pp. 1--10, 2018.

\bibitem{geert2008}
Geert Willems, Tinne Tuytelaars, and Luc~Van Gool,
\newblock ``{An Efficient Dense and Scale-Invariant Spatio-Temporal Interest
  Point Detector},''
\newblock {\em ECCV}, pp. 1--14, 2008.

\bibitem{herbert2006}
Herbert Bay, Tinne Tuytelaars, and Luc~Van Gool,
\newblock ``{SURF: Speed Up Robust Features},''
\newblock {\em ECCV}, pp. 1--14, 2006.

\bibitem{navneet2006}
Navneet Dalal, Bill Triggs, and Cordelia Schmid,
\newblock ``{Human Detection Using Oriented Histogram of Flow and
  Appearance},''
\newblock {\em ECCV}, pp. 428--441, 2006.

\bibitem{chen2018}
Mingliang Chen, Xing Wei, Qingxiong Yang, Qing Li, Gang Wang, and Ming-Hsuan
  Yang,
\newblock ``{Spatiotemporal GMM for Background Substraction with Superpixel
  Hierarchy},''
\newblock {\em TPAMI}, pp. 1518--1525, 2018.

\bibitem{mohamed2013}
Mohamed Elhoseiny, Amr Bakry, and Ahmed Elgammal,
\newblock ``{Multiclass Object Classification in Video Surveillance Systems
  Experimental Study},''
\newblock {\em CVPRW}, pp. 788--793, 2013.

\bibitem{feifei2005}
Li~Fei-Fei and Pietro Perona,
\newblock ``{A Bayesian Hierarchical Model for Learning Natural Scene
  Categories},''
\newblock {\em CVPR}, 2005.

\bibitem{kaiqi2011}
Kaiqi Huang, Dacheng Tao, Yuan Yuan, Xuelong Li, and Tieniu Tan,
\newblock ``{Biologically Inspired Features for Scene Classification in Video
  Surveillance},''
\newblock {\em IEEE Transactions on Systems, Man, and Cybernetics}, 2011.

\bibitem{Rahmani2016}
Hossein Rahmani, Arif Mahmood, Du~Huynh, and Ajmal Mian,
\newblock ``{Histogram of Oriented Principal Components for Cross-View Action
  Recognition},''
\newblock {\em TPAMI}, pp. 2430--2443, December 2016.

\bibitem{RahmaniHOPC2014}
Hossein Rahmani, Arif Mahmood, Du~Q Huynh, and Ajmal Mian,
\newblock ``{HOPC: Histogram of Oriented Principal Components of 3D Pointclouds
  for Action Recognition},''
\newblock in {\em ECCV}, 2014, pp. 742--757.

\bibitem{mitko2009}
Mitko Veta, Tomislav Kartalov, and Zoran Ivanovski,
\newblock ``{Content-based Indoor/Outdoor Video Classification System for a
  Mobile Platform},''
\newblock {\em International Journal of Electrical and Computer Engineering},
  2009.

\bibitem{limin2018}
Limin Wang, Wei Li, Wen Li, and Luc~Van Gool,
\newblock ``{Appearance-and-Relation Networks for Video Classification},''
\newblock {\em CVPR}, 2018.

\bibitem{haichen2017}
Haichen Shen, Seungyeop Han, Matthai Philipose, and Arvind Krishnamurthy,
\newblock ``{Fast Video Classification via Adaptive Cascading of Deep
  Models},''
\newblock {\em CVPR}, 2017.

\bibitem{xiang2018}
Xiang Long, Chuang Gan, Gerard de~Melo, Jiajun Wu, Xiao Liu, and Shilei Wen,
\newblock ``{Attention Clusters: Purely Attention Based Local Feature
  Integration for Video Classification},''
\newblock {\em CVPR}, 2018.

\bibitem{du2015}
Du~Tran, Lubomir Bourdev, Rob Fergus, Lorenzo Torresani, and Manohar Paluri,
\newblock ``{Learning Spatiotemporal Features with 3D Convolutional
  Networks},''
\newblock {\em ICCV}, pp. 4489--4497, 2015.

\bibitem{joao2018}
Joao Carreira and Andrew Zisserman,
\newblock ``{Quo Vadis, Action Recognition? A New Model and the Kinetics
  Dataset},''
\newblock {\em CVPR}, pp. 1--10, 2018.

\bibitem{katsunori2016}
Katsunori Ohnishi, Masatoshi Hidaka, and Tatsuya Harada,
\newblock ``{Improved Dense Trajectory with Cross Streams},''
\newblock {\em ACMMM}, pp. 1--6, 2016.

\bibitem{limin2015}
Limin Wang, Yu~Qiao, and Xiaoou Tang,
\newblock ``{Action Recognition with Trajectory-Pooled Deep-Convolutional
  Descriptors},''
\newblock {\em CVPR}, pp. 1--10, 2015.

\bibitem{kensho2017}
Kensho Hara, Hirokatsu Kataoka, and Yutaka Satoh,
\newblock ``{Learning Spatio-Temporal Features with 3D Residual Networks for
  Action Recognition},''
\newblock {\em ICCV}, pp. 3154--3160, 2017.

\bibitem{uijlings2014}
J.R.R. Uijlings, I.C. Duta, N.~Rostamzadeh, and N.~Sebe,
\newblock ``{Realtime Video Classification using Dense HOF/HOG},''
\newblock {\em ICMR}, 2014.

\bibitem{heng2011}
Heng Wang, Alexander Klaser, Cordelia Schmid, and Liu Cheng-Lin,
\newblock ``{Action Recognition by Dense Trajectories},''
\newblock {\em CVPR}, pp. 3169--3176, 2011.

\bibitem{alexander2008}
Alexander Klaser, Marcin Marszalek, and Cordelia Schmid,
\newblock ``{A Spatio-Temporal Descriptor Based on 3D-Gradients},''
\newblock {\em BMCV}, pp. 1--10, 2008.

\bibitem{paul2007}
Paul Scovanner, Saad Ali, and Mubarak Shah,
\newblock ``{A 3-Dimentional SIFT Descriptor and its Application to Action
  Recognition},''
\newblock {\em CRCV}, pp. 1--4, 2007.

\bibitem{heng2013}
Heng Wang, Alexander Klaser, Cordelia Schmid, and Cheng-Lin Liu,
\newblock ``{Dense Trajectories and Motion Boundary Descriptors for Action
  Recognition},''
\newblock {\em IJCV}, 2013.

\bibitem{heng2013iccv}
Heng Wang and Cordelia Schmid,
\newblock ``{Action Recognition with Improved Trajectories},''
\newblock {\em ICCV}, pp. 3551--3558, 2013.

\bibitem{li2015}
Li~Liu, Mengyang Yu, and Ling Shao,
\newblock ``{Unsupervised Local Feature Hashing for Image Similarity Search},''
\newblock {\em IEEE Transactions on Cybernetics}, pp. 1--11, 2015.

\bibitem{wang2017}
Jingdong Wang, Ting Zhang, Jingkuan Song, Nicu Sebe, and Heng~Tao Shen,
\newblock ``{A Survey on Learning to Hash},''
\newblock {\em TPAMI}, pp. 1--21, 2017.

\bibitem{arockiam2012}
L.~Arockiam and V.~Arul Kumar,
\newblock ``{Enhanced Feature Selection Algorithm using Modified Fisher
  Criterion and Principal Feature Analysis},''
\newblock {\em International Journal of Advanced Research in Computer Science},
  pp. 310--314, 2012.

\bibitem{sa2007}
Sa~Wang, Cheng-Lin Liu, and Lian Zheng,
\newblock ``{Feature Selection By Combining Fisher Criterion and Principal
  Feature Analysis},''
\newblock {\em International Conference on Machine Learning and Cybernetics},
  pp. 1149--1154, 2007.

\bibitem{mcinnes2018umap-software}
Leland McInnes, John Healy, Nathaniel Saul, and Lukas Grossberger,
\newblock ``Umap: Uniform manifold approximation and projection,''
\newblock {\em The Journal of Open Source Software}, vol. 3, no. 29, pp. 861,
  2018.

\bibitem{2018arXivUMAP}
L.~{McInnes} and J.~{Healy},
\newblock ``{UMAP: Uniform Manifold Approximation and Projection for Dimension
  Reduction},''
\newblock {\em ArXiv e-prints}, Feb. 2018.

\bibitem{florent2007}
Florent Perronnin and Christopher Dance,
\newblock ``{Fisher Kernels on Visual Vocabularies for Image Categorization},''
\newblock {\em CVPR}, pp. 1--8, 2009.

\bibitem{florent2010}
Florent Perronnin, Jorge Sanchez, and Thomas Mensink,
\newblock ``{Improving the Fisher Kernel for Large-Scale Image
  Classification},''
\newblock {\em ECCV}, pp. 143--156, 2010.

\end{thebibliography}

\end{document}